\documentclass[times, review, 10pt]{elsarticle}

\usepackage{amssymb}
\usepackage{amsmath}
\usepackage{algorithm}
\usepackage{algorithmic}
\usepackage{pifont}
\usepackage{natbib}  %
\usepackage{caption} %
\usepackage{newfloat}
\usepackage{listings}
\usepackage{amsmath}
\usepackage{natbib}
\usepackage{graphicx}
\usepackage{booktabs}

\newcommand{\myMethod}{\textsc{Sda-Planner}}

\newcommand{\stateGraph}{State-Dependency Graph Generation}
\newcommand{\errorBD}{Error Backtrack and Diagnosis}
\newcommand{\adaTree}{Adaptive Action SubTree Generation}

\begin{document}

\begin{frontmatter}

\title{\myMethod: State-Dependency Aware Adaptive Planner for Embodied Task Planning}

\author[label1]{Zichao Shen} %
\ead{shenzc@buaa.edu.cn}

\author[label1]{Chen Gao\corref{cor1}}
\ead{gaochen.ai@gmail.com}

\author[label1]{Jiaqi Yuan}
\ead{yuanjqqq@gmail.com}
\author[label1]{Tianchen Zhu}
\ead{zhutc@act.buaa.edu.cn}

\author[label2]{Xingcheng Fu}
\ead{fuxc@gxnu.edu.cn}

\author[label1]{Qingyun Sun}
\ead{sunqy@buaa.edu.cn}
\affiliation[label1]{organization={Beihang University},%
            city={Beijing}, 
            postcode={100191},
            country={China}}

\affiliation[label2]{organization={Guangxi Normal University},%
            city={Guangxi}, 
            postcode={541004},
            country={China}}
\cortext[cor1]{Corresponding authors}

\begin{abstract}
Embodied task planning requires agents to produce executable actions in a close-loop manner within the environment. With progressively improving capabilities of LLMs in task decomposition, planning, and generalization, current embodied task planning methods adopt LLM-based architecture.
   However, existing LLM-based planners remain limited in three aspects, \emph{i.e.}, fixed planning paradigms, lack of action sequence constraints, and error-agnostic.
   In this work, we propose \myMethod, enabling an adaptive planning paradigm, state-dependency aware and error-aware mechanisms for comprehensive embodied task planning. 
   Specifically, \myMethod~introduces a State-Dependency Graph to explicitly model action preconditions and effects, guiding the dynamic revision.
   To handle execution error, it employs an error-adaptive replanning strategy consisting of Error Backtrack and Diagnosis and Adaptive Action SubTree Generation, which locally reconstructs the affected portion of the plan based on the current environment state. 
   Experiments demonstrate that \myMethod~consistently outperforms baselines in success rate and goal completion, particularly under diverse error conditions.
\end{abstract}


\end{frontmatter}

\section{Introduction}
Embodied intelligence entails the capacity of agents to perceive, interpret, and act within the environments~\citep{FANG2021107822,liu2024aligning,CHEN2026112177,gao2023room,LIU2025111448}. 
A crucial ability of the embodied agent is task planning, \textit{i.e.}, decomposing high-level natural language instructions into coherent sequences of intermediate, goal-directed, mid-level actions~\citep{kaelbling2011hierarchical}.
In recent years, Large Language Models (LLMs)~\citep{LLMsurvey,grattafiori2024llama,yang2025qwen3,liu2024deepseek,jiang2024mixtral,glm2024chatglm} have demonstrated strong generalization capabilities across a wide spectrum of tasks. 
Owing to their ability to encode rich, implicit knowledge about the world\citep{zhang2023large, roberts2020much}, LLMs have emerged as promising components of embodied agents, particularly for task planning~\citep{huang2024understanding}.

\begin{figure*}
    \centering
    \includegraphics[width=\linewidth]{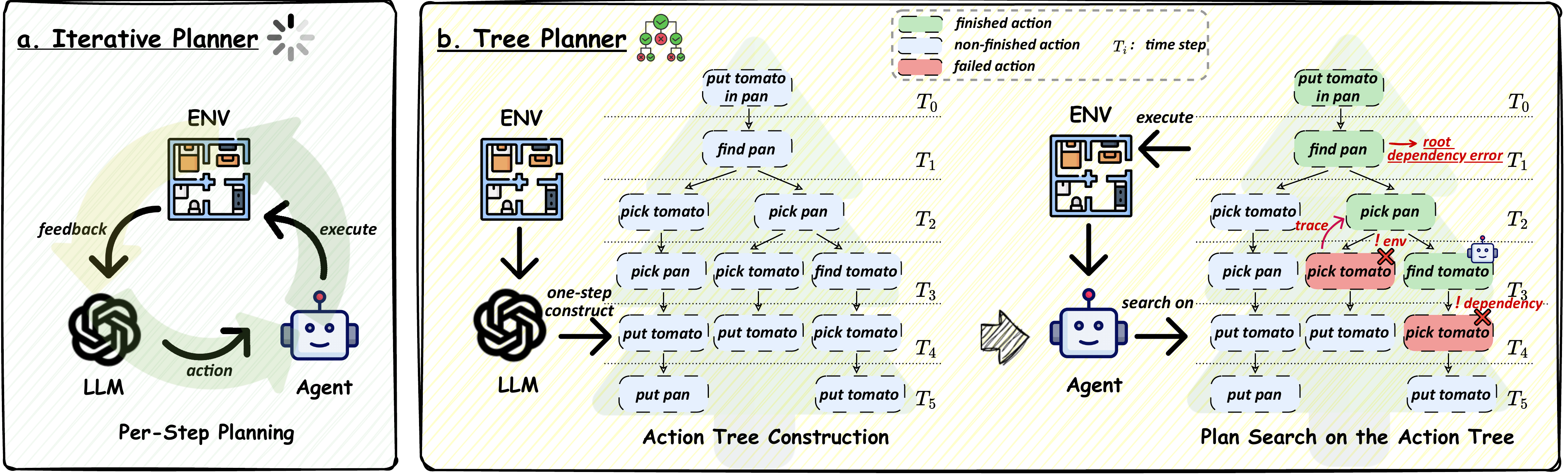}
    \vspace{-1.2em}
    \caption{Paradigm overview of existing methods. (a) Iterative Planner produces an action at every time step. (b) Tree Planner generates the action tree in advance and conducts the plan search without adjusting the original tree.}
    \label{fig:planner}
\end{figure*}

Existing LLM-based task planners for embodied agents generally fall into two categories (shown in Fig.\ref{fig:planner}): 
(1) \textit{Iterative Planner}, 
which generates one action at a time based on real-time feedback from the environment, 
and (2) \textit{Tree Planner}~\citep{hu2023tree}, 
which generates entire static plans using an LLM and constructs an action tree for execution.
Despite their promise, these paradigms exhibit several notable limitations:
\ding{182} \textbf{Fixed Planning Paradigms}: 
Iterative Planners prompt an LLM to generate one action at a time, which heavily relies on repeated interactions between LLMs and the environment. Such a redundant and fixed paradigm results in high time and token costs, making the process inefficient.
Conversely, Tree Planner prompts the LLM only once at the initial stage to generate the entire candidate plan, \textit{i.e.}, the action tree. Then, it constrains subsequent trajectory search within the fixed structure of the action tree, limiting adaptability to new information or errors during execution.
\ding{183} \textbf{Lack action sequence constraints}: 
Both the Iterative and Tree Planners typically treat actions as isolated steps, without explicitly modeling dependencies between them. 
As a result, agents may attempt actions like ``\textit{place tomato}'' without satisfying preconditions such as ``\textit{pick up tomato}''. Iterative Planner reacts step-by-step but lacks a global view to enforce such constraints, while Tree Planner reasons on the generated action tree and cannot make targeted adaptations if earlier actions invalidate later steps.
\ding{184} \textbf{Error-agnostic}: 
The encountered errors during execution can be broadly categorized into two types:
(1) \textit{Environment State Errors}, which stem from a mismatch between the agent’s assumptions and the actual environment state. 
(2) \textit{Action Precondition Errors}, which arise due to internal flaws in the plan itself and often result from implicit or unmodeled dependencies between actions~\citep{li2024embodied}. 
Such errors typically reflect missing or violated preconditions that were not properly accounted for during the original planning phase.
Neither planner can adjust the plan according to different errors. 
For example, if the action ``\textit{pick up tomato}'' fails, Iterative Planner repeatedly attempts without re-evaluating. Tree Planner can only search for another path in the fixed action tree. Thus, both planners are error-agnostic and cannot detect/correct the root error.

To overcome these limitations, we propose \textbf{S}tate-\textbf{D}ependency \ Aware \textbf{A}daptive \textbf{Planner} (\textbf{\myMethod}), a novel framework designed to enable error-aware and adaptive embodied task planning. 
\myMethod~is built upon three key components:
First, the \textbf{State-Dependency Graph Generation} module explicitly models action-state dependencies, enforcing structural constraints to ensure actions in the reconstructed subsequence are only executed when their preconditions are satisfied.
Second, the \textbf{Error Backtrack and Diagnosis} module enables structured and targeted error handling by identifying root causes (\textit{e.g.}, unmet preconditions) and isolating the minimal action subsequence requiring revision, avoiding full-plan regeneration.
Third, the \textbf{Adaptive Action SubTree Generation} module reconstructs the affected subsequence using current environment context and constraints from the dependency graph, enabling efficient and localized replanning.
Together, these components enable \myMethod~to reason over complex state-action dependencies, differentiate between diverse error types, and dynamically adjust plans at a fine-grained level. 
We conduct experiments on the ALFRED benchmark and \myMethod~achieves a superior success rate and goal-condition success rate compared to existing fixed-paradigm and error-agnostic planners, highlighting its adaptability in embodied task planning.
Our contributions are as follows:
\begin{itemize}
\item 
We propose \myMethod, a novel planning framework for embodied agents that integrates state dependency modeling with error-aware replanning, enabling robust task execution in environments.
\item 
We design a state-aware and error-specific replanning mechanism that leverages a state-dependency graph for modeling action preconditions, and performs localized replan through error-aware diagnosis and adaptive action subtree generation.
\item 
Extensive experiments demonstrate that \myMethod~outperforms strong baselines in both task success and goal completion under various execution error scenarios.
\end{itemize}

\section{Related Work}
\subsection{Planning with LLMs}
Numerous methods~\citep{gao2023adaptive,gao2021room,gao2025octonav,wu2023embodied,singh2022progprompt,dagan2023dynamic,wang2024llmˆ} aim to achieve embodied task planning. Recent methods adopt LLMs and can be divided into two categories based on how LLMs are involved in the planning process. The first category~\citep{ahn2022can,guan2023leveraging} uses LLMs to \textit{indirectly} generate plans, such as ranking optional skills or generating intermediate representations, and the intermediate results are further processed to obtain the final plan. But these methods often lead to inefficiencies. The other category~\citep{song2023llm, shin2025socraticplannerselfqabasedzeroshot} uses LLMs to \textit{directly} generate plans. This category of methods often provides LLMs with relevant context examples to guide the generation of plan. However, these methods lack explicit modeling of action state dependencies and constraints on action sequences, which often leads to the generated plan not being able to be effectively executed in the environment.

\subsection{LLMs Re-planning}
Although LLM has great potential for planning, it still faces problems such as hallucinations and environment mismatches. To solve this problem, one category of methods uses prompt engineering to enhance the planning capability of LLM~\citep{zhu2023ghost}. Another category of methods uses multi-agent collaboration techniques to enhance LLM planning capability by employing multiple agents to collaborate with each other~\citep{qin2024mp5,wang2023describe}. In addition to this, when errors are encountered during the execution of the plan, the researchers used replanning techniques to correct the plan so that the task could be completed successfully~\citep{raman2022planning}. 
According to the scope of plan regeneration, the related methods can be categorized into local replan~\citep{guo2024doremi} and global replan~\citep{shinn2023reflexion}.
Specifically, local replan regenerates the plan from the current time step. This method ensures efficiency, but it is difficult to resolve errors that occurred before the current timestep. In contrast, global replan generates the entire plan from scratch, but it tends to suffer from inefficiency and non-retroactivity of some actions.

\section{Preliminary}
Formally, the task planning process based on Large Language Models (LLMs) can be defined as follows: 
given a natural language instruction $\mathcal{I}$ and a predefined set of executable skills $\mathcal{V}$, the LLM is responsible for decomposing $\mathcal{I}$ into a sequence of mid-level actions $\mathcal{P}$, defined as:
\begin{equation}\label{eq:p}
    \mathcal{P} =\{\mathcal{A}_0,\cdots,\mathcal{A}_t\}= \{(a_0, o_0),\cdots,(a_t, o_t)\},
\end{equation}
where $a_t\in \mathcal{V}$ denotes an action type, $o_t$ represents the target object of the action, and $(a_t,o_t)$ constitutes a specific mid-level action denoted as $\mathcal{A}_t$.
Subsequently, each $\mathcal{A}$ is translated by the low-level planner into a primitive action to interact with the environment. 
For example, for the instruction ``\textit{put the bread on the table}'', a possible mid-level action sequence would be \{(``\textit{find}'', ``\textit{bread}''), (``\textit{pick up}'', ``\textit{bread}''), (``\textit{find}'', ``\textit{table}''), (``\textit{put down}'', ``\textit{bread}'')\}.

\section{Method}

\subsection{Overview of \myMethod}
\begin{figure*}[!htp]
    \centering
    \includegraphics[width=\linewidth]{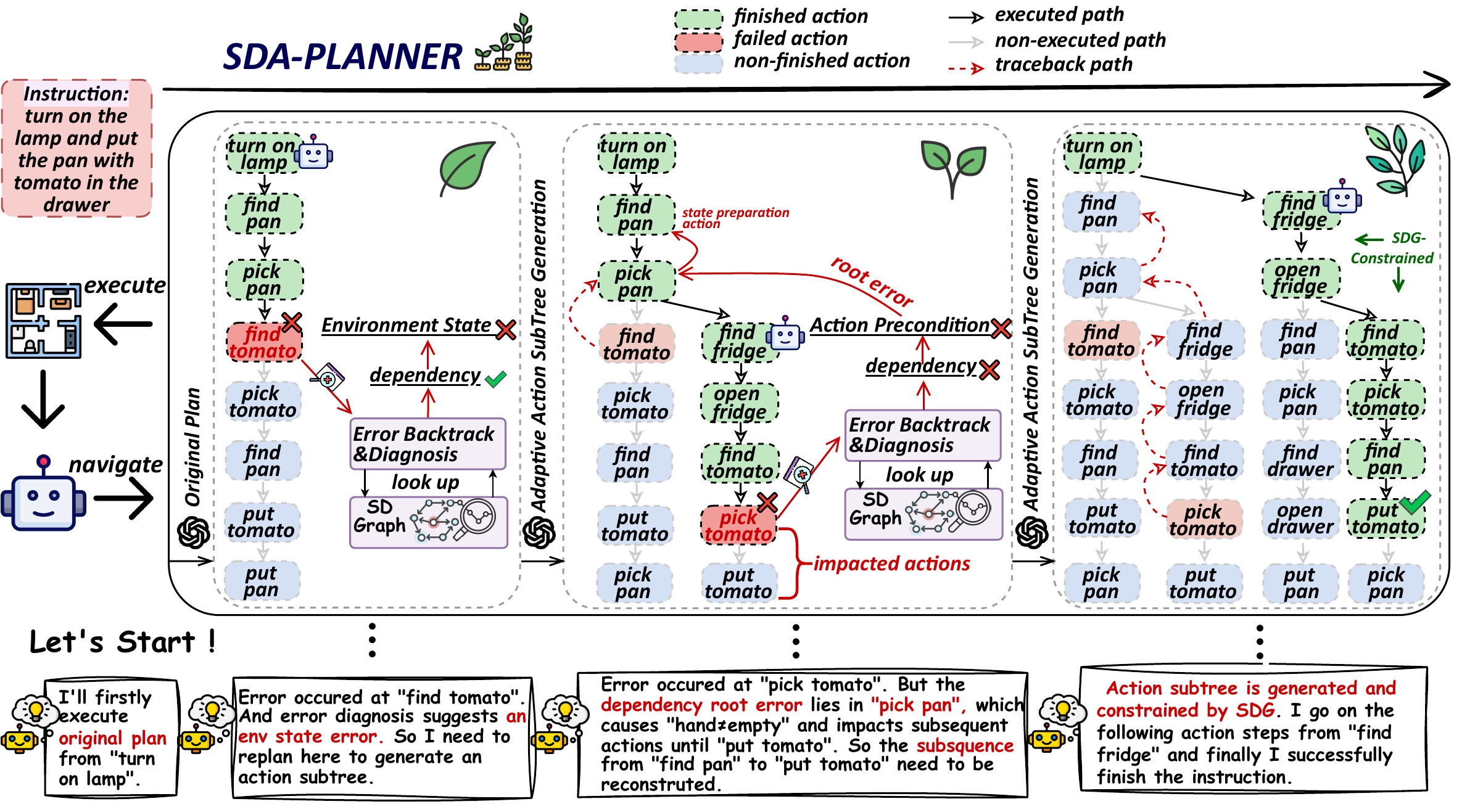}
\vspace{-1.2em}
    \caption{Overview of \myMethod. When executing the instruction
    , \myMethod~first breaks down the plan and later adapts the plan when it encounters an error.
    \myMethod~analyses with the state dependency graph and dynamically adapts the plan by different methods based on the error type,  to solve the environment state error and the action precondition error.
    }
    \label{fig:framework}
\end{figure*}

To address the challenges faced by existing LLM-based planners,
we propose a novel framework (shown in Fig.~\ref{fig:framework}), \myMethod, which integrates three tightly coupled components to enable state-dependency aware and adaptive task planning in dynamic environments. 
Specifically, \myMethod~comprises the \stateGraph~module for dependency modeling, the \errorBD~module for error diagnosis and targeted plan adjustment, and the \adaTree~module for adaptive plan reconstruction.

\textit{\stateGraph}~module ($\rhd$~\textit{Section~\ref{sec:sdg}}) constructs a state-dependency graph that explicitly captures the preconditions and effects of mid-level actions. 
This graph provides structural constraints that guide the subsequent replanning, ensuring that each action in the subsequence is only executed when its required state conditions are satisfied. 

If an execution failure occurs, \textit{\errorBD}~module ($\rhd$~\textit{Section~\ref{sec:diagnosis}}) is activated to diagnose the error based on the structure of the dependency graph. 
For errors caused by environmental dynamics (\textit{e.g.}, unexpected object positions or occlusions), the module adopts a local replanning strategy to adjust the plan. In contrast, for errors arising from violated action preconditions (\textit{e.g.}, attempting to pick up an object with an already occupied hand), the module identifies the minimal invalid subsequence that must be reconstructed.

Building on this diagnosis, \textit{\adaTree}~module ($\rhd$~\textit{Section~\ref{sec:adatree}}) generates a revised plan subtree that satisfies the updated state constraints. 
It does so by constructing a constrained search tree, which incorporates the error information and dependency structure to produce context-aware, minimal plan modifications. 
To ensure consistency between the updated plan and the physical state of the environment, \myMethod~incorporates a backtracking mechanism that reverses previously executed actions when necessary. 
Furthermore, it adopts a \textit{fake execution} strategy to simulate the impact of planned actions before executing, thereby identifying and mitigating potential conflicts in advance.

\subsection{State-Dependency Graph Generation}
\label{sec:sdg}
\stateGraph~module constructs a State-Dependency Graph $\mathcal{G}$ to explicitly capture the preconditions and effects of mid-level actions. 
Unlike prior methods that rely on implicit LLM reasoning with its internal knowledge, this structured representation enhances interpretability and ensures consistency in adaptive replanning.

Formally, given the agent’s skill set $\mathcal{V}$, for each action $a \in \mathcal{V}$, we define two associated state sets: effect set $\mathcal{S}_\text{eff}[a]$ and dependency set $\mathcal{S}_\text{dep}[a]$. 
Each state in these sets is represented as a pair consisting of a state variable and its expected value. 
The state variables are further categorized into agent states and item states, which are dynamically grounded to specific entities when the action is instantiated as a mid-level action $\mathcal{A}_i = (a_i, o_i)$.
The effect set $\mathcal{S}_\text{eff}[a]$ denotes the set of environment or agent states changed as a consequence of executing $a$.
The dependence set $\mathcal{S}_\text{dep}[a]$ describes the states that must be satisfied prior to executing $a$ successfully.
For example, for the action ``\textit{pick up}'', its effect set $\mathcal{S}_\text{eff}[a]$ contains the handheld state of the agent (changing the handheld state to the corresponding object). 
And its dependence set $\mathcal{S}_\text{dep}[a]$ contains the agent's positional state (which requires the agent to be physically close to the object before attempting the action). 

To automatically construct the two state sets for each action in $\mathcal{V}$, 
\stateGraph~module leverages a hybrid approach: it can either query an LLM for commonsense-driven annotations or rely on external task-specific knowledge bases. 
The process for LLM-based construction is divided into three steps.  
First, during initialization, a predefined set of base states is used to bootstrap the reasoning process. Then, for each action $a$, the module queries the LLM in a structured prompt format to get the $\mathcal{S}_\text{eff}[a]$ and expands the state set. Finally, based on the state set, the module queries the LLM for $\mathcal{S}_\text{dep}[a]$. This automated extraction not only improves scalability but also reduces reliance on manual action schemas.

\stateGraph~module then constructs the state-dependency graph $\mathcal{G}$ using two state sets $\mathcal{S}_\text{eff}$ and $\mathcal{S}_\text{dep}$. 
$\mathcal{G}$ is modeled as a directed bipartite graph, where the node set is partitioned into two disjoint subsets:
The first subset $N_a$ contains action nodes $n_a$, each corresponding to an action $a\in \mathcal{V}$.
The second subset $\mathcal{N}_s$ contains state nodes $n_s$, each representing a state variable $s$ that appears in either the action's effect set $\mathcal{S}_\text{eff}$ and $\mathcal{S}_\text{dep}$.
Edges in the graph are directed and encode the relationship between actions and states:
A directed edge from an action node $n_a$ to a state node $n_s$ indicates that action $a$ modifies the state $s$. The edge is annotated with a value $v$, denoting the postcondition, \textit{i.e.}, the new value that $s$ takes after executing $a$.
Conversely, a directed edge from a state node $n_s$ to an action node $n_a$ signifies that the action $a$ requires the state $s$ to have value $v$ as a precondition for successful execution.
In task execution, when applying the State Dependency Graph to a specific $\mathcal{A}_i = (a_i, o_i)$, it is only necessary to refine the names and values of the related state nodes and edges to the corresponding $o_i$.

Moreover, we can categorize certain actions as \textit{state preparation action} using the state dependency graph. 
Formally, an action $a$ is considered a state preparation action if its corresponding node $n_a$ has exactly one outgoing edge to an agent state node
and no incoming edges from other state nodes. 
These actions are not dependent on any prior state and serve to initialize or adjust the agent’s state to satisfy the preconditions of subsequent actions.
An example is ``\textit{find}''. This action is a state preparation action, which is used to adjust the positional state of the agent in preparation for later actions such as ``\textit{pick up}'', without relying on other states.

By capturing the structural dependencies in this way, the State Dependency Graph serves as a foundational abstraction that informs error diagnosis, action sequencing, and context-aware replanning throughout the planning pipeline.
\subsection{Error Backtrack and Diagnosis}
\label{sec:diagnosis}
When an error is encountered, \myMethod~first recognizes the type of error with the help of \errorBD~module.
In the case of action precondition errors, it further localizes the subsequence of reconstructed actions.

Let the mid-level action sequence be denoted as $\mathcal{P}=\{(a_1, o_1), (a_2, o_2), ...\}$, an error occurs at timestep $t_\text{error}$ and the corresponding action-object pair is $(a_\text{error},o_\text{error})$. 
First, \errorBD~module analyses the dependent states of $(a_\text{error}, o_\text{error})$ sequentially, checking whether each of them is satisfied based on the executed actions $(a_t, o_t)$.
If all dependent states are satisfied, the module uses the local replan strategy to generate additional action steps from the current time step onward.
Otherwise, the module further localizes the subsequence that needs to be modified based on $t_\text{error}$ and the unsatisfied state dependency $s_\text{error}$. Then, the subsequence is reconstructed by the \adaTree~module.

To formalize this, we denote the desired value of the unsatisfied state $s_\text{error}$ as $v_\text{need}$,  and define the start and end positions of the subsequence to be reconstructed as $t_\text{start}$ and $t_\text{end}$, respectively.
If the state node $n_s$ corresponding to $s_\text{error}$ has only one incoming edge in $\mathcal{G}$ and this edge originates from a state preparation action node $n_a$, this implies that the error stems solely from the agent's internal state being unprepared.
In this case, the plan can be adapted by directly inserting the corresponding state preparation action at timestep $t_\text{error}$, without the full reconstruction.
At this point, we set $t_\text{start}=t_\text{error}=t_\text{end}$. 
However, if the state dependency involves more complex factors, such as object-related conditions or interactions between multiple entities, then a deeper revision is required. 
In such cases, we first define the error source point $t_\text{source}$, which is the most recent time step prior to $t_\text{error}$ at which the required state $s_\text{error}$ changed from being satisfied to unsatisfied:
\begin{equation}\label{eq:t_source}
    t_{\text{source}} = 
    \left\{
    \begin{aligned}
        & \operatorname*{max} \{t | t \in \Lambda\}, &\quad \Lambda \neq \emptyset \\ 
        & 1, & \quad \Lambda = \emptyset ,
    \end{aligned}
    \right.
\end{equation}
where $\Lambda$ is defined as the set of all corrupted moments for the target state dependency: 
\begin{equation}\label{eq:lambda}
\small
    \begin{aligned}
    \Lambda = \{t | t< t_\text{{error}} \land (s_\text{error}[t-1] = v_\text{need}) \land (s_\text{error}[t] \neq v_\text{need})\}.
    \end{aligned}
\end{equation}
We then define the reconstruction window $[t_\text{start},t_\text{end}]$ based on $t_\text{source}$ and $t_\text{error}$ as follows:
\begin{equation}
\small
\begin{aligned}
        t_{\text{start}} &= \min\{  t \mid 
         \{ a_i \mid \mathcal{A}_i = (a_i, o_i), 
         \forall i \in [t, t_{\text{source}}) \} \subseteq \mathcal{A}_{\text{prep}}\}
         \\
     t_{\text{end}} &= 
    \max\{  t \mid 
         \{ o_i \mid A_i = (a_i, o_i), 
         \forall i \in (t_{\text{error}}, t] \} \subseteq \mathcal{O}\},
\end{aligned}
\end{equation}
where $\mathcal{A}_\text{prep}$ is the set consisting of state preparation actions in $\mathcal{V}$, and $\mathcal{O}$ denotes the set of error items, which contains $o_\text{error}$ as well as the corresponding items in $s_\text{error}$.

\begin{figure}
    \centering
    \includegraphics[width=\linewidth]{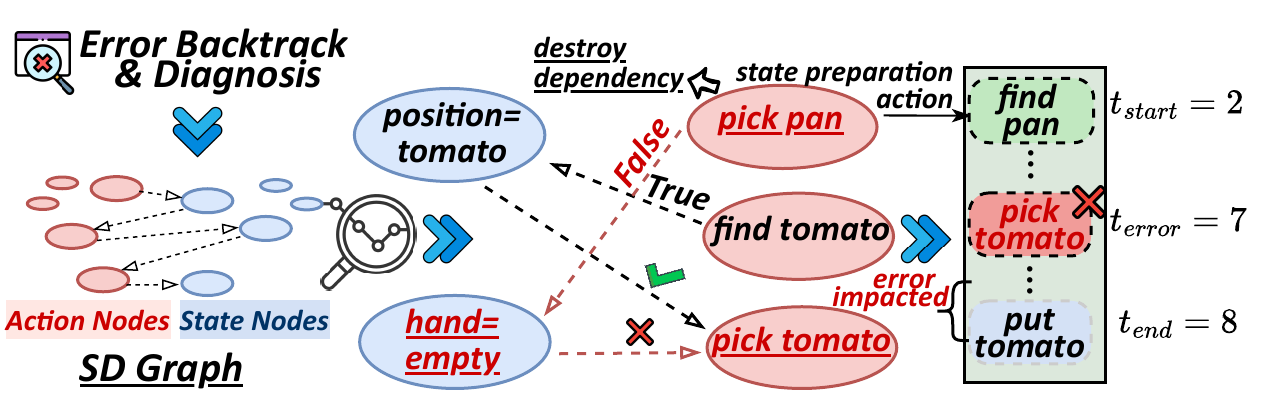}
    \caption{Illustration of the \errorBD.
    }
    \label{fig:grpah}
\end{figure}
As illustrated in Fig.~\ref{fig:framework}, 
consider a failure case where the agent attempts (``\textit{pick up}'', ``\textit{tomato}'') at $t=7$, but the action fails. The \errorBD~module then analyses the error with the state dependency graph as shown in Fig.~\ref{fig:grpah}.
The root cause is traced to (``\textit{pick up}'', ``\textit{pan}'') at $t=3$, which altered the holding state and violated the precondition of holding nothing.
Thus, $t_\text{error}=7$ and $t_\text{source}=3$.
The preceding (``\textit{find}'', ``\textit{pan}'') at $t=2$ is a preparatory action, setting $t_\text{start}=2$, and since both the pan and tomato are affected by this error, $t_\text{end}=8$.
This example highlights \myMethod’s ability to localize and revise the minimal necessary plan subsequence to recover task feasibility.

\subsection{Adaptive Action SubTree Generation}
\label{sec:adatree}
After identifying the subsequence to be revised, \adaTree~module constructs a search tree to ensure that the reconstructed sequence satisfies all action preconditions.
Each node represents a mid-level action $\mathcal{A}=(a, o)$, and each path from the root denotes a candidate executable plan.
The reconstruction proceeds in three stages: (1) generating nodes, (2) building the tree, and (3) extracting a valid plan from it.

In the first stage, \adaTree~module utilizes an LLM to analyze the causes of error in the original subsequence,  and LLM further outputs the relevant corrective actions based on the relevant information provided.
The nodes of the search tree are then generated using two sources: the corrective actions recommended by the LLM, and the actions in the original subsequence. 
In particular, the actions provided by LLM can make this state dependency error effectively corrected, while the actions provided by the original subsequence can ensure that all relevant actions in the original execution plan are fully taken into account. 
To reduce the size of the subsequent search space, we further introduce a constraint on the reconstruction process. 
Specifically, if a subsequence in the original plan $\mathcal{P}'=\{\mathcal{A}_i, \cdots,\mathcal{A}_j\}$ satisfies the condition that all involved items are the same, \textit{i.e.}, $o_i=o_{i+1}=\cdots=o_n$, and $o_n \notin O$, then we infer that this subsequence was designed to exert a complete and uninterrupted influence on $o_n$ and should not be split during the reconstruction process.
Accordingly, we designate actions $\mathcal{A}_{i+1}$ to $\mathcal{A}_j$ as non-selectable nodes in the search tree, while treating $\mathcal{A}_i$ as a special optional node.

Subsequently, \adaTree~module constructs a search tree using the relevant candidate nodes. The action at the root node of the tree is defined as a special empty node, whose associated state is set to the state corresponding to the node $\mathcal{A}_{\rm start-1}$, \textit{i.e.}, the step immediately preceding the error start point.
This root node serves as the starting point for generating corrective action sequences.
We then define how child nodes are constructed for each node. 
For the previously mentioned subsequence $\mathcal{P}'= \{\mathcal{A}_i, \cdots,\mathcal{A}_{j-1}$\}, if node $\mathcal{A}_t$ belongs to this subsequence ($\mathcal{A}_t \in \mathcal{P}'$), then its child node is fixed as $\mathcal{A}_{t+1}$. 
For other nodes not in such constrained subsequences, the set of child nodes $N$ for $\mathcal{A}_t$ is defined by:
\begin{equation}\label{eq:child}
\small
    \begin{aligned}
        N = \{\mathcal{A}_j \in \{\mathcal{V}_r - \mathcal{V}_{\text{used}}\}  \mid \text{satisfied}(\mathcal{A}_j, \mathcal{G}) \\
    \land \text{change}(\mathcal{A}_j, \mathcal{G}) \land \text{notCovered}(\mathcal{A}_t, \mathcal{A}_j)\},
    \end{aligned}
\end{equation}
where $\mathcal{V}_r$ is the list of optional nodes derived from the actions provided by LLM and the actions in the original subsequence, and $\mathcal{V}_{used}$ is the list of optional nodes in the sequence of actions corresponding to $\mathcal{A}_t$. 
$\text{satisfied}(\mathcal{A}_j, \mathcal{G})$ and $\text{change}(\mathcal{A}_j, \mathcal{G})$ denote that $\mathcal{A}_j$ satisfies the constraints of the state dependency graph $\mathcal{G}$ and can have an effect on the current state.
$\text{notCovered}(\mathcal{A}_t, \mathcal{A}_j)$ is defined as:
\begin{equation}\label{eq:cover}
\small
    \text{notCovered}(\mathcal{A}_t, \mathcal{A}_j) = 
    \left\{
    \begin{aligned}
        & \text{True}, &\exists s, \begin{aligned}
            (\mathcal{A}_t, s)\in \mathcal{E} \land (\mathcal{A}_j, s) \notin \mathcal{E}
        \end{aligned} \\ 
        & \text{False}, & otherwise,
    \end{aligned}
    \right.
\end{equation}
where $\mathcal{E}$ denotes the set of edges of $\mathcal{G}$. The child node should not override the $S_\text{eff}$ of the parent node. 
After constructing the search tree, the \adaTree~module further considers the state constraints and performs a breadth-first search to extract a fully executable subsequence. 
Subsequently the derived subsequence is combined with the remaining original steps as the action subtree of the original plan. 
Finally, to guide planning, both the environment context and task instructions are provided, enabling the LLM to evaluate alternatives and choose an optimal, task-aligned plan from the tree.

To ensure consistency between the revised plan and the environment state, after \adaTree~module completes its adaptation process, \myMethod~first performs a reverse execution of the actions between $t_\text{start}$ and $t_\text{error}$.
This reversal aims to restore the environment state as closely as possible to its condition at $t_\text{start}$ (\textit{e.g.}, executing a ``\textit{pick up}'' action in response to a prior ``\textit{put down}''). 
After reversal, \myMethod~re-executes the adapted plan from $t_\text{start}$ to carry out the remaining steps under the revised strategy.
In particular, for \textit{irrecoverable actions} that cannot restore the original state, it adopts a \textit{fake execution} strategy, \textit{i.e.}, skipping previously executed irreversible actions to prevent state conflicts.

In the example of Fig.~\ref{fig:framework}, 
after plan adaptation, \myMethod~first executes the reverse actions (``\textit{close}'', ``\textit{fridge}'') and (``\textit{put down}'', ``\textit{pan}'')  corresponding to the reversal of actions between $t_\text{start}=2$ and $t_\text{error}=7$ to partially restore the prior environment state. 
It then re-executes the adapted plan starting from $t_\text{start}$ to complete the task execution.
The \adaTree~module uses the search tree to ensure that the reconstructed subsequence adheres to action constraints, enabling consistent and effective recovery.

\section{Experiments}
\subsection{Experimental Setups}
\subsubsection{Datasets}
We conduct experiments on the ALFRED benchmark~\citep{shridhar2020alfred}, which requires agents to complete specified tasks in a home environment based on natural language instructions and visual inputs. 
Tasks range from simple goals like “\textit{Place the soap on the rack}” to complex multi-step instructions such as “\textit{Slice the bread and chill it in the fridge}”, requiring strong reasoning and adaptability.

\subsubsection{Baselines and Settings}
We choose two categories of baselines: (1) non-LLM-generated methods and (2) LLM-based planners.
For the non-LLM-generated baselines, we include \textit{HLSM}~\citep{blukis2022persistent} and \textit{Saycan}~\citep{ahn2022can}.
We report results directly from related papers.
For LLM-based planners, we selected \textit{LLM planner (Local Planner)}, \textit{Global Planner}, and \textit{Tree Planner}~\citep{hu2023tree}. 
\textit{LLM Planner}~\citep{song2023llm} directly generates plans via LLMs and injects the observed objects into the prompt to adapt the plan. 
For the original \textit{LLM Planner}, we combined the local replan approach to modify it. 
\textit{Global Planner} adopts the global replan strategy. When it encounters errors, \textit{Global Planner} regenerates the complete plan and executes it from scratch. 
Note that after generating the plan, we manually reset the relevant environment, which is unrealistic.
\textit{Tree Planner} builds an action tree via sampling and performs replanning over the tree. 
All relevant methods are evaluated under the unified LOTA-BENCH~\citep{choi2024lota} framework, using consistent prompts, low-level controllers, and the GPT-4o-mini~\citep{achiam2023gpt} backend.
Experiments are conducted on a server with Intel Xeon Gold 6148 CPU @ 2.40GHz. 
We fix the random seed to 1 and set the LLM temperature to 0 for deterministic outputs.%

\subsubsection{Metrics}
We choose three metrics: Success Rate (SR), Goal Condition Success Rate (GC), and Number of Error Corrections (No. EC).
SR is the percentage of tasks in which all sub-goals are successfully completed.
GC is the proportion of sub-goals completed per task.
No. EC is the average number of plan corrections triggered per task, measuring the adaptability to errors.
Because \textit{Tree Planner}'s replanning process relies on the action tree, we only count No.EC of \textit{LLM Planner}, \textit{Global Planner}, and \myMethod.

\begin{table*}[!htp]
  \centering
  \caption{Main results on ALFRED, where the best results are shown in \textbf{bold} and the runner-ups are shown in underlined.}
  \label{exp-table}
  \resizebox{\textwidth}{!}{
  \begin{tabular}{ll*{9}{c}}
    \toprule
    \multicolumn{1}{c}{}&\multicolumn{1}{c}{}&\multicolumn{3}{c}{Valid Seen}& \multicolumn{3}{c}{Valid Unseen} & \multicolumn{3}{c}{Average}       \\
    \cmidrule(r){3-5} \cmidrule(r){6-8} \cmidrule(r){9-11}
    Method&  LLM  & SR & GC & No.EC & SR & GC & No.EC & SR & GC & No.EC\\
    \midrule
    HLSM& - &29.63&38.74&-&18.28&31.24&-&23.96&34.99&-\\
    SayCan& GPT-3  &12.30&24.52&-&9.88&22.54&-&11.09&23.53&-\\
    LLM planner& GPT-4o-mini   & 31.32 &40.97&3.75 & 40.00 &51.15&3.51&35.66&46.06&3.63 \\
    Global Planner& GPT-4o-mini  & \underline{35.92} &  \underline{42.72} &  2.42 & \underline{42.70}  &\underline{51.22}&2.06&\underline{39.31}&\underline{46.97}&2.24\\
    Tree Planner& GPT-4o-mini     & 32.47    &  39.82 & - &40.54 &48.36&-&36.51&44.09&-\\
    \myMethod & GPT-4o-mini           & \textbf{38.22}  &  \textbf{47.10} &3.01 &\textbf{44.32} &\textbf{54.73}&3.10&\textbf{41.27}&\textbf{50.92}&3.06\\
    \bottomrule
  \end{tabular}
}
\end{table*}

\subsection{Main Results}
As shown in Table~\ref{exp-table},
planners using GPT-4o-mini outperform the training-based baseline \textit{HLSM} in both SR and GC, highlighting the effectiveness of LLM-based planning. 
Among all methods, \myMethod~achieves the best overall performance, with SR and GC improved by 2\% and 4\%, respectively, demonstrating its ability to generate high-quality, executable plans.
In terms of error correction, \myMethod~requires only 3 corrections per task on average, significantly fewer than the \textit{LLM Planner(Local Planner)}, which reflects the strength of its adaptive error-handling mechanism. 
Although \textit{Global Planner} has the lowest No. EC, this is attributed to its inefficiency: it restarts execution from scratch upon error, which reduces the number of corrections but often fails to recover from errors effectively.
\begin{figure}[!tp]
\centering
\includegraphics[width=0.99\linewidth]{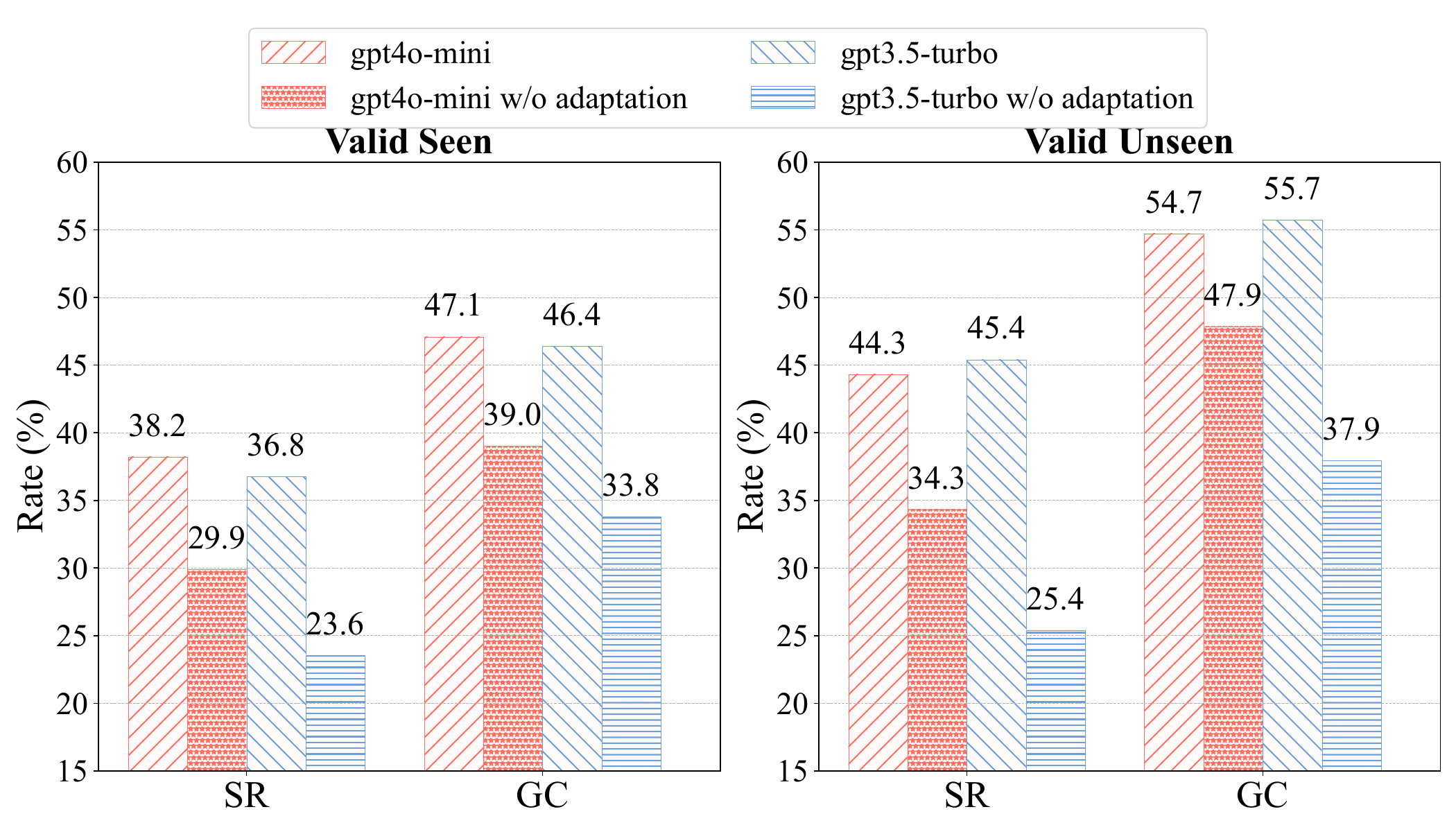}
\caption{Ablation study with different LLMs.}
\label{fig:bar_two}
\end{figure}
\subsection{Ablation Study}
To assess the contribution of each component in \myMethod, we conduct ablation studies focusing on (1) the choice of LLM and (2) the effect of plan adaptation. 

First, we replace GPT-4o-mini with GPT-3.5-Turbo in the task decomposition and plan adaptation stages, and we also disable the plan adaptation function of \myMethod. In this setting, when an error is encountered, \myMethod~no longer adapts the plan, but skips the error and continues to execute the plan. We denote this variant as ``\textit{w/o adaptation}.'' Results are shown in Fig.~\ref{fig:bar_two}.
Results highlight the importance of plan adaptation, which significantly improves SR and GC. The search tree in \adaTree~further enhances adaptation by enforcing structural constraints during reconstruction.
Notably, GPT-3.5-Turbo shows greater performance degradation without adaptation, highlighting that plan adaptation is especially vital for less capable LLMs.

\begin{figure}[!tp]
    \centering
\includegraphics[width=0.99\linewidth]{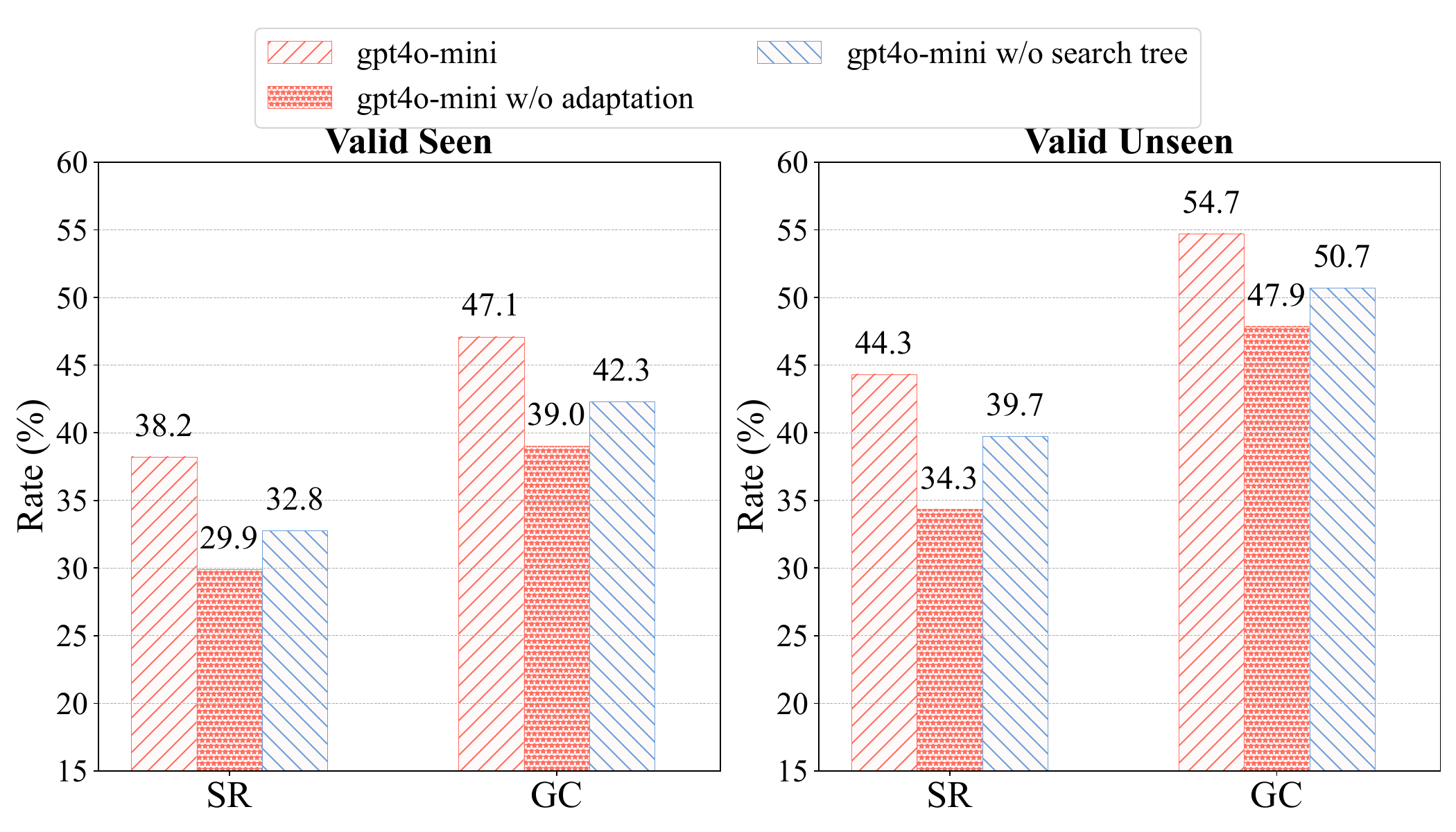}
    \caption{Ablation study with different adaptation methods.}
    \label{fig:bar_fill}
\end{figure}
We further study the reconstruction strategy in \adaTree. 
Instead of using a search tree as hard constraints, we adopt a soft constraint variant denoted as ``\textit{w/o search tree}'' that provides violated preconditions as prompts to the LLM for direct subsequence generation.
Results in Fig.~\ref{fig:bar_fill} show that the variant with prompt constraint leads to reduced SR and GC, indicating that 
\adaTree~module benefits from combining LLMs’ semantic reasoning with structural constraints from the search tree, leading to more reliable plans.
\begin{figure*}[!ht]
    \centering
    \includegraphics[width=\linewidth]{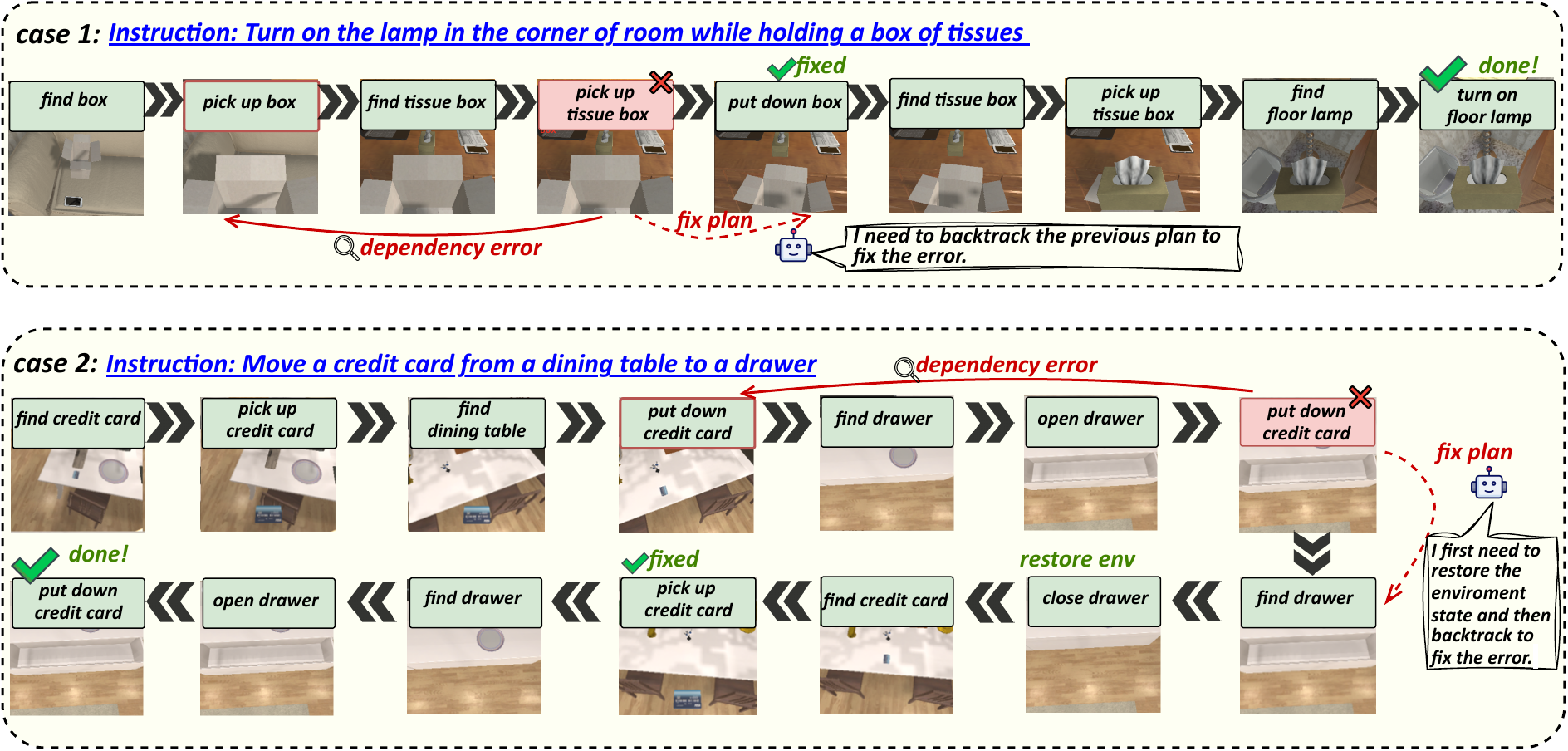}
    \caption{Case study for \myMethod.}
    \label{fig:case}
\end{figure*}
\subsection{Case Study}
We show two cases in Fig.~\ref{fig:case}. 
In case 1, during the execution of the task  ``\textit{turn on the lamp in the corner of the room while holding a box of tissues}'', the agent fails at (``\textit{pick up}'', ``\textit{tissue box}''), due to an earlier (``\textit{pick up}'', \textit{box}'') at $t=2$.
So \myMethod~starts reconstructing the subsequence from (``\textit{find}'', ``\textit{box}'') and executes (``\textit{put down}, ``\textit{box}'') to restore the state of the environment.
For case 2, during the task ``\textit{Move a credit card from a dining table to a drawer}'', the action (``\textit{put down}'', ``\textit{credit card}'') at $t=7$ fails. Then \myMethod~uses the \errorBD~module to analyse the error.
\errorBD~module traces the issue to $t=4$, where the agent lost the holding state for the credit card, and then reconstructs the subsequence starting from (``\textit{find}'', ``\textit{dining table}''). 
After that, \myMethod~starts to execute the reverse actions, (``\textit{close}'', ``\textit{drawer}'') and (``\textit{pick up}'', ``\textit{credit card}''), and continues to execute the corrected plan after the reversal is complete. 
These two cases highlight \myMethod's capability for precise error localization, state restoration, and adaptive plan correction.

\section{Conclusion}
In this work, we presented \myMethod, a state-dependency aware adaptive planner for embodied task planning. 
By integrating the State-Dependency Graph, Error Backtrack and Diagnosis, and Adaptive Action SubTree Generation, \myMethod~jointly address the core limitations of existing LLM-based planners including fixed planning paradigms, lack of precondition reasoning, and error-agnostic replan mechanisms.
Experimental results demonstrate that \myMethod~significantly improves success rate and goal completion rate, offering a promising step toward more reliable and generalizable planning systems for embodied intelligence.

\bibliographystyle{elsarticle-num}
\bibliography{ref}
\end{document}